# Compressed representation of brain genetic transcription

James K. Ruffle FRCR MSc[1], Henry Watkins PhD[1], Robert J Gray PhD[1], Harpreet Hyare FRCR PhD[1], Michel Thiebaut de Schotten PhD, HDR, CNRS[2,3], and Parashkev Nachev FRCP PhD[1]

[1]*Queen Square Institute of Neurology, University College London, London, UK*

[2]*Groupe d'Imagerie Neurofonctionnelle, Institut des Maladies Neurodégénératives-UMR 5293, CNRS, CEA, University of Bordeaux, France*

[3]*Brain Connectivity and Behaviour Laboratory, Paris, France*

Correspondence to:

Dr James K Ruffle

Email: j.ruffle@ucl.ac.uk

Address: Institute of Neurology, UCL, London WC1N 3BG, UK

Correspondence may also be addressed to:

Professor Parashkev Nachev

Email: p.nachev@ucl.ac.uk

Address: Institute of Neurology, UCL, London WC1N 3BG, UK

## Funding

JKR was supported by the Medical Research Council (MR/X00046X/1), the NHS Topol Digital Fellowship and the UCL CDT i4health. PN is supported by the Wellcome Trust (213038/Z/18/Z) and the UCLH NIHR Biomedical Research Centre. HH is supported by the UCLH NIHR Biomedical Research Centre. MTS received funding from the European Research Council (ERC) under the European Union's Horizon 2020 research and innovation programme (Grant Agreement No. 818521) and conducted his work within the framework of the University of Bordeaux's IHU 'Precision & Global Vascular Brain Health Institute - VBHI' and the RRI "IMPACT", which received financial support from the France 2030 program. Data were provided by the Human Connectome Project, WU-Minn Consortium (Principal Investigators: David Van Essen and Kamil Ugurbil; 1U54MH091657) funded by the 16 NIH Institutes and Centres that support the NIH Blueprint for Neuroscience Research and by the McDonnell Center for Systems Neuroscience at Washington University.

## Conflict of interest

None to declare.

## Manuscript Type

Technical report.

## Authorship

Conceptualisation: JR, PN; Data: JR, MTS; Methodology: JR, HW, RJG, MTS, PN; Software: JR, HW, RJG; Validation: JR, HW, RJG; Formal analysis: JR; Manuscript writing JR, MTS, PN; Manuscript reviewing, and editing: JR, HW, RJG, HH, MTS, PN. All authors have been involved in the writing of the manuscript and have read and approved the final version.

## Abbreviations

PCA, principal component analysis; t-distributed stochastic neighbour embedding (t-SNE); uniform manifold approximation and projection (UMAP).

# Abstract

The architecture of the brain is too complex to be intuitively surveyable without the use of compressed representations that project its variation into a compact, navigable space. The task is especially challenging with high-dimensional data, such as gene expression, where the joint complexity of anatomical and transcriptional patterns demands maximum compression. The established practice is to use standard principal component analysis (PCA), whose computational felicity is offset by limited expressivity, especially at great compression ratios. Employing whole-brain, voxel-wise Allen Brain Atlas transcription data, here we systematically compare compressed representations based on the most widely supported linear and non-linear methods—PCA, kernel PCA, non-negative matrix factorisation (NMF), t-stochastic neighbour embedding (t-SNE), uniform manifold approximation and projection (UMAP), and deep auto-encoding—quantifying reconstruction fidelity, anatomical coherence, and predictive utility across signalling, microstructural, and metabolic targets, drawn from large-scale open-source MRI and PET data. We show that deep auto-encoders yield superior representations across all metrics of performance and target domains, supporting their use as the reference standard for representing transcription patterns in the human brain.

# Introduction

There is increasing recognition of the need to model the brain's anatomy within richly expressive computational frameworks capable of capturing complex, spatially distributed neural organisation. A network perspective on the brain requires explicit representation of the relations between functionally critical regions as well as their identity, revealing structure jointly conveyed in the interactions of multiple anatomical loci[1-3]. The combinatorial explosion this entails, even at coarse anatomical scales, demands compressed latent representations that maximise descriptive fidelity while minimising dimensionality.

This concern is not limited to distributed anatomy. It also applies to the surfeit of high-dimensional characteristics—such as microstructure, metabolism, signalling, and genetic expression—by which individual anatomical loci may be differentiated[4,5]. The challenge of obtaining faithful compressed representations is especially great with genetic expression, where high dimensionality is coupled with potentially complex long-range interactions[6,7]. The established approach has been to use principal component analysis (PCA)[8] to reduce transcriptive variation at each anatomical locus to a single latent dimension[9-11]. Though mathematically appealing, PCA has been shown to leave most of the transcriptomic variance unexplained here, capturing only 20-30% of the total in the first two dimensions[9,12]. PCA moreover models the data in purely linear terms that is both implausible and contradicted by transcription patterns elsewhere[13-17].

Recent advances in representation learning offer an array of alternative approaches. Choosing between them rests on the balance of fidelity, compactness, organisation of the latent representation, and—crucially—its utility in downstream tasks such as outcome prediction and spatial inference. In a comprehensive analysis of Allen Brain Atlas transcription data[18], here we systematically train and evaluate a set of candidate models in the task of obtaining compressed representations of large-scale transcription data across the whole human brain. For each model—spanning standard PCA[8], kernel PCA[19], nonnegative matrix factorisation (NMF)[20], t-distributed stochastic neighbour embedding (t-SNE)[21], uniform manifold approximation and projection (UMAP)[22], and deep auto-encoding[23,24]—we quantify reconstruction fidelity on held-out test data, characterise the latent structure, and establish utility in downstream predictive performance across 24 physiological targets encompassing regional neurotransmitter distributions, synaptic density, myelination, and metabolic activity, drawing upon large-scale open-source MRI and PET data. Our analysis demonstrates that deep auto-encoders yield substantially superior representations across all performance metrics and supports their use as the reference standard for representing transcription patterns in the human brain.

## Methods

### Data

All available microarray transcription data across the whole human brain was derived from all available donors (n=6) across 3702 distinct tissue samples within the Allen Brain Atlas[5,18]. We undertook our analysis voxelwise, in both 4mm$^3$ and 8mm$^3$ resolution. A 'parcellation-free', voxel-wise approach minimises the risk of spatial bias arising from data-driven parcellation schemes[25-31] whose organisation may unpredictably interact with transcription patterns. To quantify generalisability across different resolutions, we replicated all models at both 4mm$^3$ and 8mm$^3$ voxel resolution, equating to 29 298 and 3670 brain voxels, respectively. This voxel-wise approach provides finer anatomical detail than most published parcellation schemes, as detailed elsewhere[3].

### Transcription processing pipeline

We used the Abagen toolbox[5] to retrieve Allen Brain Atlas microarray gene expression across the whole brain[18], as described in detail elsewhere[5,32]. In brief, this included, i) intensity-based filtering to remove probes that do not exceed background noise (set to the standard value of 0.5); ii) selection of the optimal probe for each gene (set to the standard option of 'diff_stability', i.e. selecting the probe with the most consistent pattern of regional variation across donors); and iii) mirroring of tissue samples across left/right hemispheric boundaries in instances where samples are acquired only unilaterally, in accordance with established practice[5]. Microarray samples were assigned to each voxel by first determining whether the sample was located within the given voxel, and where not, expanding the search space to identify nearby probes. This process yielded a set of 15 633 transcription data for every voxel. At each voxel, we clamped the 15 633 microarray gene expression profiles between the 0.1$^{th}$ and 99.9$^{th}$ percentiles to minimise the impact of outliers. Data were already scaled within the 0-1 range.

### Model training and testing

We randomly split voxels into 80% training and 20% test sets. All representation learning techniques detailed below were initialised and fitted to the training set, with subsequent application of the fitted transform to the test set. The specific voxels included in the train and test partitions were the same for all techniques to ensure the only difference in embedding result would be due to the method used.

### Representation learning

We considered the range of available methods for representation learning of complex data, including those specifically applicable to neuroimaging and transcriptomics. The methods employed, chosen on the grounds of popularity, theoretical foundations, and utility in kindred domains, were the following: 1) principal component analysis (PCA)[8]; 2) RBF kernel PCA[19]; 3)

NMF[20]; 4) t-SNE, with automatic learning rate and perplexity value of 30[21]; 5) UMAP, with 30 nearest neighbours and a minimum distance of 0.1[22]; and 6) an auto-encoder (AE). Each method was used to embed whole brain transcription data to 2, 4, 8, 16, 32, 64, and 128 dimensions, from both 4mm³ and 8mm³ resolutions, to quantify performance as a function of representational dimensionality. PCA is an algorithm for representing multivariate data by finding orthogonal linear combinations of the features such that each dimension in turn explains as much of the variance as possible. kPCA uses a "kernel trick" to apply the PCA algorithm to a simple non-linear transformation of the feature vectors[19]. NMF is an algorithm for factorising a matrix into non-negative, more easily interpretable matrices. T-SNE and UMAP are nonlinear dimension reduction techniques commonly used for visualising high-dimensional data owing to their ability to disentangle clusters in high-dimensional spaces. The former aligns two models of the joint probabilities of data points based on distance, one in the original space and one in a low-dimensional space[21]. The latter pursues the same objective within a framework based on Riemannian geometry and algebraic topology[22]. Parameter choices are wholly guided by the authors of the algorithms and/or software APIs as referenced above.

The auto-encoder uses a standard fully connected architecture, minimising inductive bias. We use batch normalisation to reduce vulnerability to suboptimal parameter initialisation and to enable higher learning rates[33]. We use exponential linear unit activation functions based on their efficacy in other contexts[23,24,33-36]. The input dimensionality is 15 633 (the length of the source data vector for each voxel). The encoder layer sizes, in order, were 500, 250, 125, then 2. The decoder layer sizes are the same but ordered in reverse. The output layer of the decoder uses a sigmoid activation function, bounding the output in the range [0,1]. The number of parameters in the model was 15 966 885. Model hyperparameter tuning was accomplished by 5-fold cross-validation within the training set. The Adam optimiser was used for its exceptional rate of convergence[37]. The learning rate was 0.001, and the batch size was 128. The loss function is mean squared error, plus $L_2$ regularisation to shrink weights towards zero to minimise overfitting[38]. Each model was trained for a maximum of 50 000 epochs with early stopping from 300 epochs, also to minimise overfitting.

### Quantifying representational fidelity

Representational fidelity was quantified by computing the reconstruction error (root mean squared error (RMSE)) of test data decoded from each representation. Given the clearly Gaussian distribution of source data (Supplementary Figure 1), the RMSE loss function was deemed most appropriate. This was possible for PCA, kPCA, NMF, UMAP, and the AE, but not t-SNE, where no inverse transform implementation is available. Owing to the computational difficulty of data reconstruction in UMAP (see discussion here[39]), only 2-component representations at 4mm³ and 8mm³, and only 4-component representations at 8mm³, could be evaluated under the constraint of parallelised code within a 250Gb RAM budget.

## Quantifying predictive fidelity

We subsequently quantified the predictive performance of the PCA, kPCA, NMF, t-SNE UMAP, and AE transcriptomic representations in a series of downstream prediction tasks of brain physiology, spanning neurotransmission, myelination, synaptic density, and glucose metabolism.

First, we retrieved PET maps of receptor and transporter distributions across 18 neurotransmitter systems: serotonin (5-HT)$_{1A}$, 5-HT$_{2A}$, 5-HT$_4$, 5-HT$_6$, serotonin transporter (5-HTT), alpha-4 beta-2 nicotinic receptor (α$_4$β$_2$), cannabinoid (CB)$_1$, dopamine (D)$_1$, D$_2$, $^{18}$F-fluorodopa (FDOPA), γ-Aminobutyric acid type A (GABA$_A$), histamine (H)$_3$, muscarinic acetylcholine receptor (M)$_1$, metabotropic glutamate receptor 5 (mGluR$_5$), mu-opioid receptor (MOR)$_1$, noradrenaline transporter (NAT), N-methyl-D-aspartate (NMDA) receptor, and vesicular acetylcholine transporter (VAChT)[4]. Neurotransmitters were further pooled into excitatory (5-HT$_{2A}$, 5-HT$_4$, 5-HT$_6$, α$_4$β$_2$, D$_1$, mGluR$_5$, M$_1$, NMDA) and inhibitory (5-HT$_{1A}$, CB$_1$, D$_2$, GABA$_A$, H$_3$, MOR$_1$) both for prediction of overall excitatory/inhibitory properties, as well as for visualisation purposes.

Second, myelination maps expressed by the 1) T$_1$-weighted (T1w)/T$_2$-weighted (T2w) MRI ratio, and 2) fractional anisotropy (FA), averaged across adult human connectome project participants, were also modelled[28,40,41]. The T1w/T2w image was derived from the Human Connectome Project (HCP)[28,40]. In brief, images were acquired on a Siemens Skyra 3-Tesla scanner at Washington University in St. Louis, fitted with a customised body transmitter coil with a 56 cm bore size[40]. Each participant underwent an axial 3D T1-weighted imaging dataset of the entire head, consisting of 260 slices with a voxel resolution of 0.7 × 0.7 × 0.7 mm, a TE of 2.14 ms, a TR of 2400 ms, and a flip angle of 8°. Additionally, an axial 3D T2-weighted imaging dataset covering the whole head was also acquired for each participant, consisting of 260 slices with a voxel resolution of 0.7 × 0.7 × 0.7 mm, a TE of 565 ms, a TR of 3200 ms, and a variable flip angle. Advanced Normalization Tools (ANTs, http://stnava.github.io/ANTs/) were used to produce T1w and T2w templates, which utilised an iterative combination of affine and diffeomorphic deformations[42-45]. Participants were first aligned together using affine transformation, and then the templates were built from all the subjects iteratively (n=4) using diffeomorphic deformations computed using the SyN tool in ANTs. T1w/T2w maps were computed using fslmaths as part of the FSL software package (https://fsl.fmrib.ox.ac.uk)[41].

Third, we acquired open-source PET imaging of synaptic vesicle glycoprotein 2A (SV2A) radioligand, $^{11}$C-UCB-J PET, a marker of synaptic density[46]. Fourth, we acquired open-source PET imaging of $^{18}$F-FDG as a marker of metabolic activity[47].

To enable qualitative evaluation, we overlaid the voxel-wise values of each target dataset on each 2D latent representation. Quantitatively, we trained individual XGBoost models to predict each of the 24 targets from each of the 2D embeddings, a total of 144 models. All were grid-searched with 5-fold cross-validation for hyperparameter optimisation, before a final evaluation on the held-out test set.

### Code, model, and data availability

All code, model, and imaging data will be openly available upon publication at https://github.com/jamesruffle/compressed-transcriptomics. All source data is freely available from the Allen Human Brain Atlas and the Abagen toolbox[5,18].

### Software

Analyses were performed within a Python environment with the following software packages: Abagen[5], Datashader[48], Matplotlib[49], MONAI[50], Nibabel[51], Nimfa[52], NumPy[53], pandas[54], PyTorch[55], seaborn[56], scikit-learn[57], UMAP[22], XGBoost[58].

### Compute

All experiments were performed on a 64-core Linux workstation with 256Gb of RAM and an NVIDIA 3090Ti GPU.

## Results

### Representations of brain transcription data

To obtain a high-resolution, spatially unbiased representation of whole human brain microarray gene expression, we extracted data from the Allen Brain Atlas[5,18] at 4mm$^3$ and 8mm$^3$ isotropic voxel resolution. In keeping with established practice[5,32], probe data were mirrored across hemispheres and averaged across the 6 donor samples. This process yielded 15 635 gene expression profiles across 29 298 and 3670 brain voxels at 4 mm$^3$ and 8mm$^3$ resolutions, respectively.

Across separate models for each resolution, we used PCA, kPCA, NMF, t-SNE, UMAP, and deep auto-encoding to extract 2, 4, 8, 16, 32, 64, and 128 dimensional representations of the transcription data, visualising the 2D 4mm$^3$ representations as density plots (Figure 1). Qualitatively, PCA, kPCA, and NMF yielded less structured representations than t-SNE, UMAP, or the auto-encoder.

Projection of each component into brain anatomical space revealed patterns varying in their anatomical coherence (Figure 2). PCA, kPCA, and NMF broadly differentiated between the cerebellum and the rest of the brain in the first component, and (weakly for NMF) between surface and deeper regions in the second, without any regional specificity. T-SNE highlighted a dorsoventral gradient across the whole brain in the first component and a rostrocaudal one in the second. UMAP yielded a more finely granular structure, but exhibited abrupt regional variations of doubtful anatomical fidelity. The auto-encoder representation captured multiple scales of spatial organisation in an anatomically plausible manner, distributed across the two components.

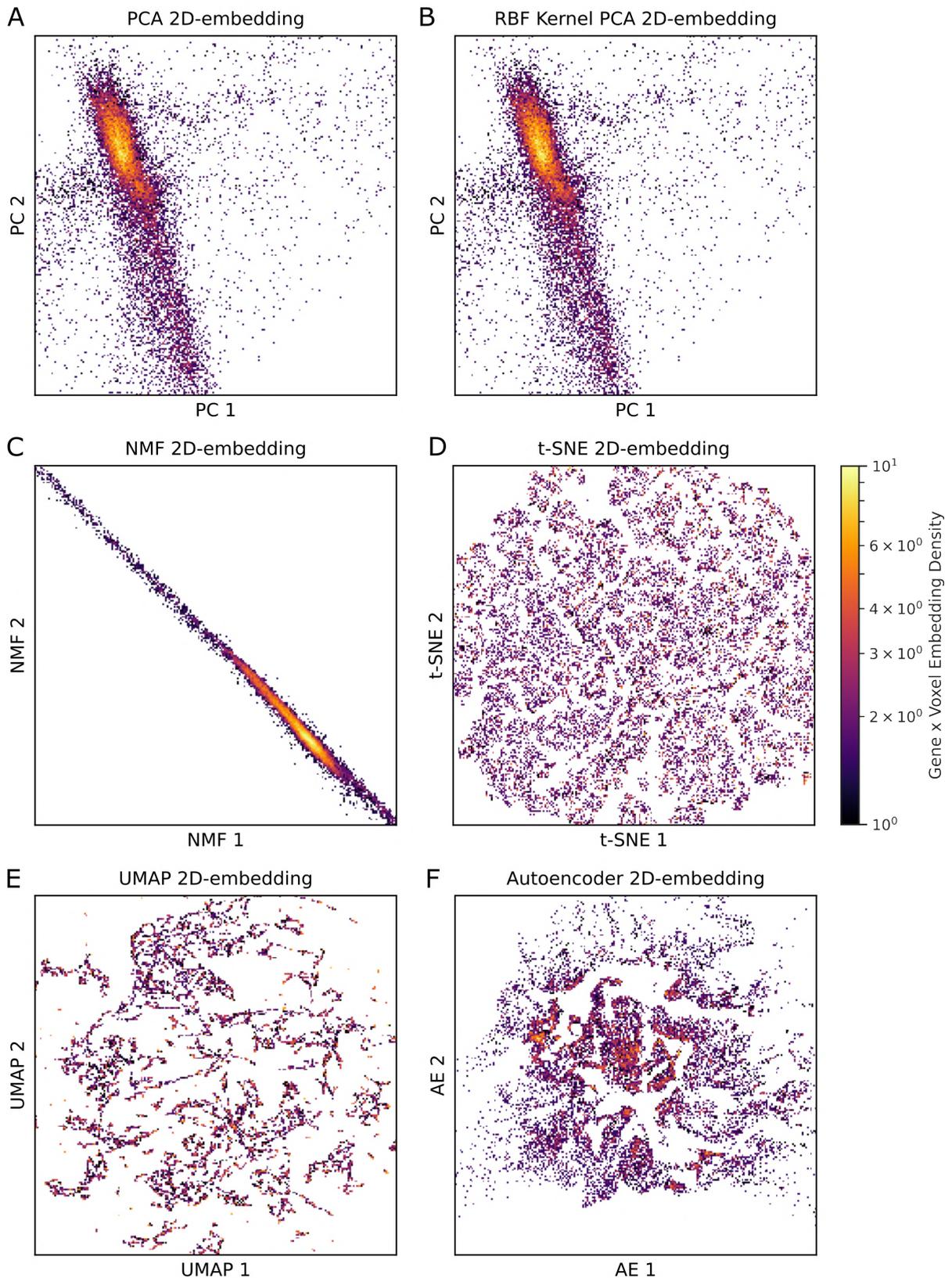

Figure 1. Density plots of 2D representations of whole brain gene expression maps based on A) PCA, B) kPCA, C) NMF, D) t-SNE, E) UMAP, and F) an auto-encoder (AE). The axes denote the first and second latent dimension of each given method. Colour is proportional to point density. Note that though conventional and kernel PCA yield visually similar latents at 2D, they differ at higher dimensions (see Figure 3).

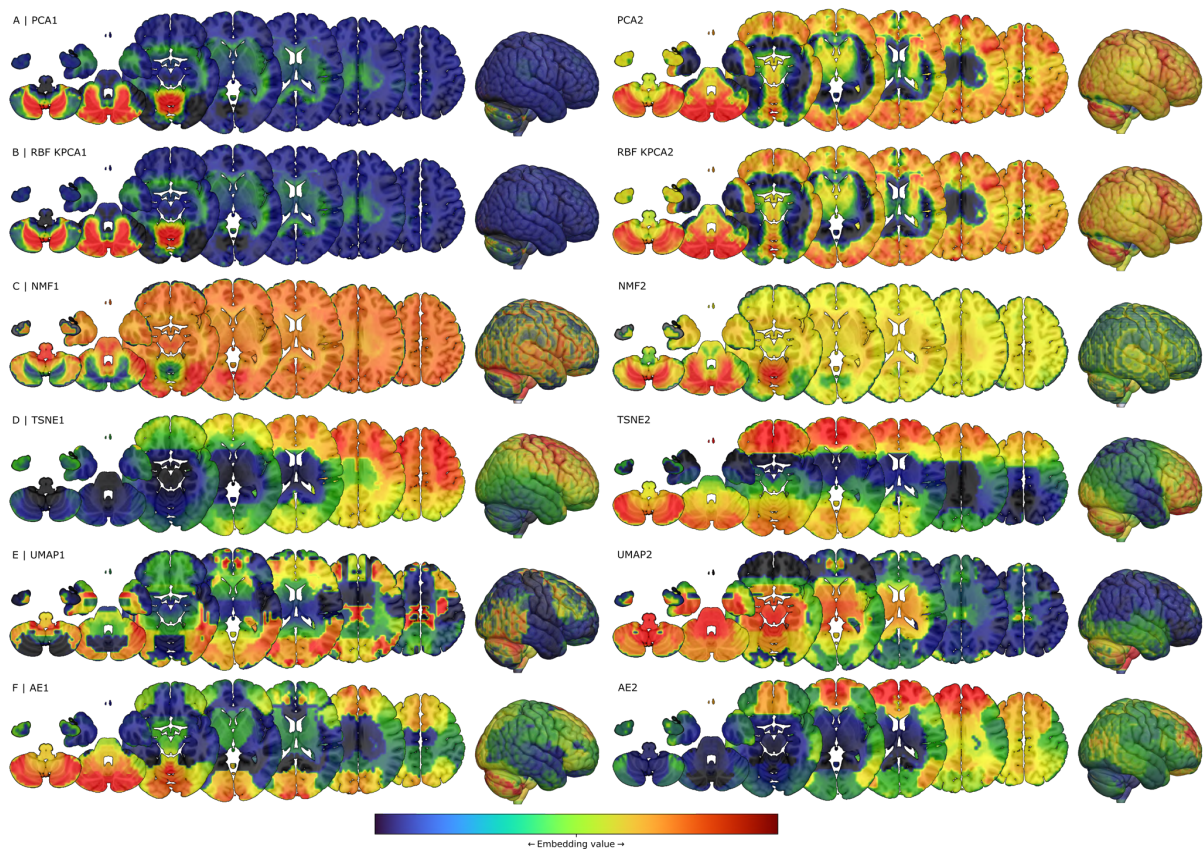

Figure 2. Brain projections of 2-dimensional genetic embeddings using A) PCA, B) radial basis function kernel PCA, C) NMF, D) t-SNE, E) UMAP, E) and F) an auto-encoder (AE). Colour is proportional to the component value at the first and second components, the units for which are arbitrary.

## Maximising reconstruction fidelity from compressed representations

Testing on held-out data, we evaluated the models' ability to reconstruct the source from representations of varying dimensionality and input image resolution (Figure 3). This was possible for all methods except for t-SNE, which does not yield an explicit function that can be applied to source reconstruction, and was limited to 2D and 4D (the latter for the 8mm$^3$ voxelate only) representations for UMAP owing to computational constraints[39]. For all input resolutions and representational dimensionalities, the auto-encoder achieved the best root-mean-squared-error (RMSE) (ANOVA p<0.0001, Tukey post-hoc comparison all p<0.0001 compared with PCA, kPCA, and NMF). Crucially, the auto-encoder representations exhibited the greatest capacity to retain fidelity in the face of compression, as indicated by the shallow slope of the relationship between RMSE and representational dimensionality.

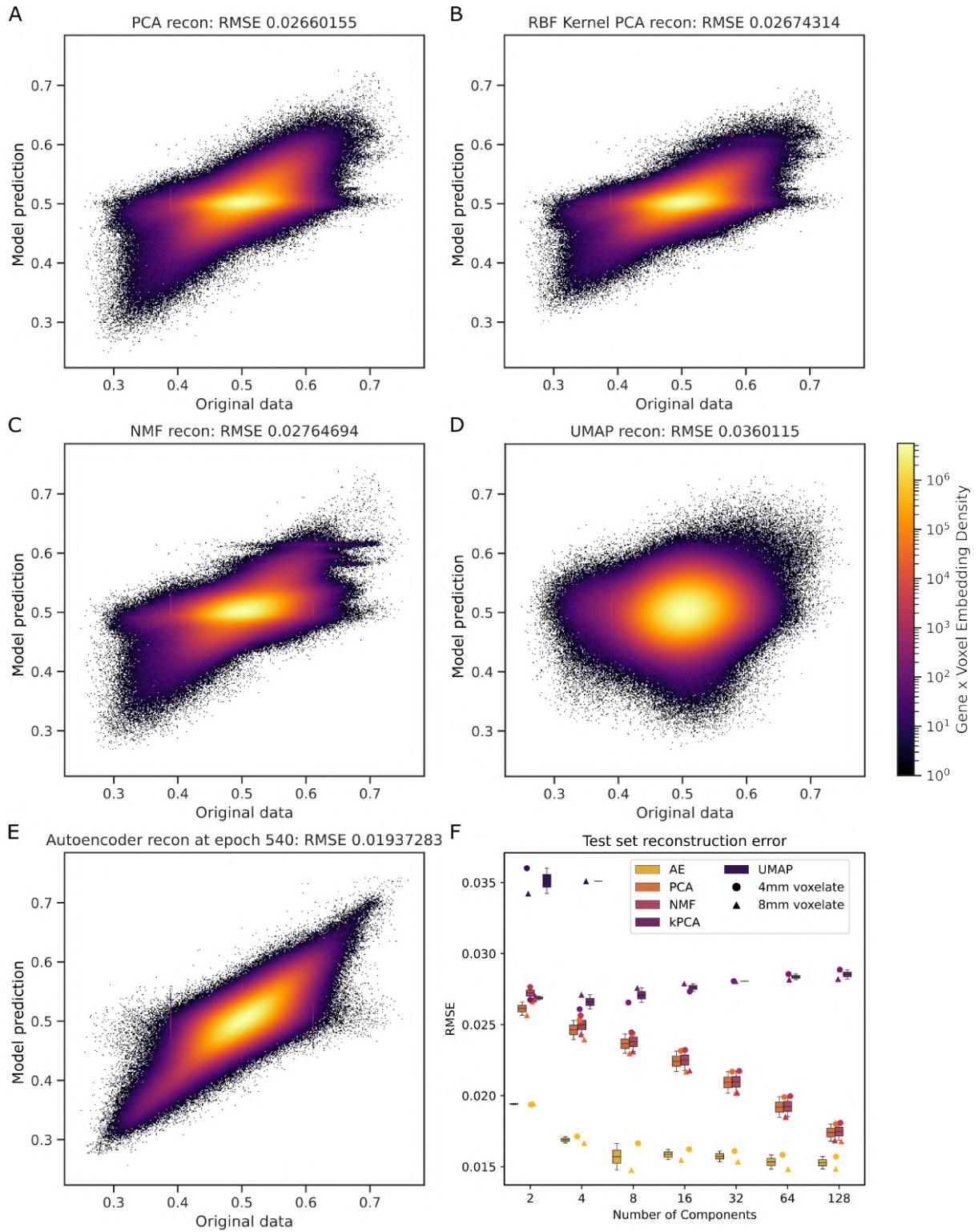

Figure 3. Test set reconstruction model performance. Reconstruction errors for A) PCA, B) kPCA, C) NMF, D) UMAP, and E) auto-encoder (AE), showing original data values on the x-axis, and reconstructed predictions on the y-axis. F) Box and whisker plot with superimposed individual points depicts reconstruction performance (by RMSE) across all component size and voxel resolutions, colour-coded by mean RMSE performance ordinally. Note, reconstruction of the original data is not possible for t-SNE owing to the nature of the algorithm, and for UMAP was computationally feasible only with 2 components at 4mm³ and 8mm³ resolutions, and 4 components at 8mm³ resolution (see discussion of technical obstacles here[39]). Note clear superiority of auto-encoding across all resolutions and dimensionalities.

## Predicting brain signalling, microstructure, and metabolism from transcriptional representations

A faithful representation of transcription patterns ought to exhibit a coherent relationship with underlying signalling, microstructural, and metabolic features. We evaluated positron emission tomography (PET) maps of receptor and transporter distributions across 18 neurotransmitter systems: serotonin (5-HT)$_{1A}$, 5-HT$_{2A}$, 5-HT$_4$, 5-HT$_6$, serotonin transporter (5-HTT), alpha-4 beta-2 nicotinic receptor (α$_4$β$_2$), cannabinoid (CB)$_1$, dopamine (D)$_1$, D$_2$, $^{18}$F-fluorodopa (FDOPA), γ-Aminobutyric acid type A (GABA$_A$), histamine (H)$_3$, muscarinic acetylcholine receptor (M)$_1$, metabotropic glutamate receptor 5 (mGluR$_5$), mu-opioid receptor (MOR)$_1$, noradrenaline transporter (NAT), N-methyl-D-aspartate (NMDA) receptor, and vesicular acetylcholine transporter (VAChT)$^4$. We also evaluated maps of myelination density captured by both T$_1$-weighted/T$_2$-weighted MRI ratio and fractional anisotropy averaged across adult human connectome project participants[28,40], PET imaging of synaptic vesicle glycoprotein 2A (SV2A) radioligand $^{11}$C-UCB-J PET as a marker of synaptic density[46], and $^{18}$F-FDG PET as a marker of metabolic activity[47].

Inspection of each 2D representation annotated by each feature revealed varying degrees of qualitative coherence (Figure 4). The most expressive representations—UMAP, t-SNE, and auto-encoding—yielded the most structured apparent relationships, with auto-encoding in particular revealing multiple scales of related organisation.

To capture the strength of the relationships quantitatively, we trained an array of XGBoost models to predict each of the 24 feature targets from each of the 2D representations, across a total of 144 individual models. All models were hyper-parameter tuned with 5-fold cross-validation and evaluated on the held-out test set (Figure 5). The auto-encoder achieved the best average RMSE and $R^2$ on the held-out test set across all experiments (mean RMSE 0.1295, mean $R^2$ 0.3563). A one-way ANOVA found a significant difference in model performance across representational methods ($p<0.0001$). Tukey post-hoc comparison showed the auto-encoder, UMAP, and t-SNE all yielded significantly superior performance (by $R^2$) than PCA, kPCA, and NMF (all $p<0.0001$). There was no significant difference between auto-encoder, UMAP, and t-SNE performance ($p=0.866$ or higher), or between PCA, kPCA, and NMF ($p=0.555$ or higher).

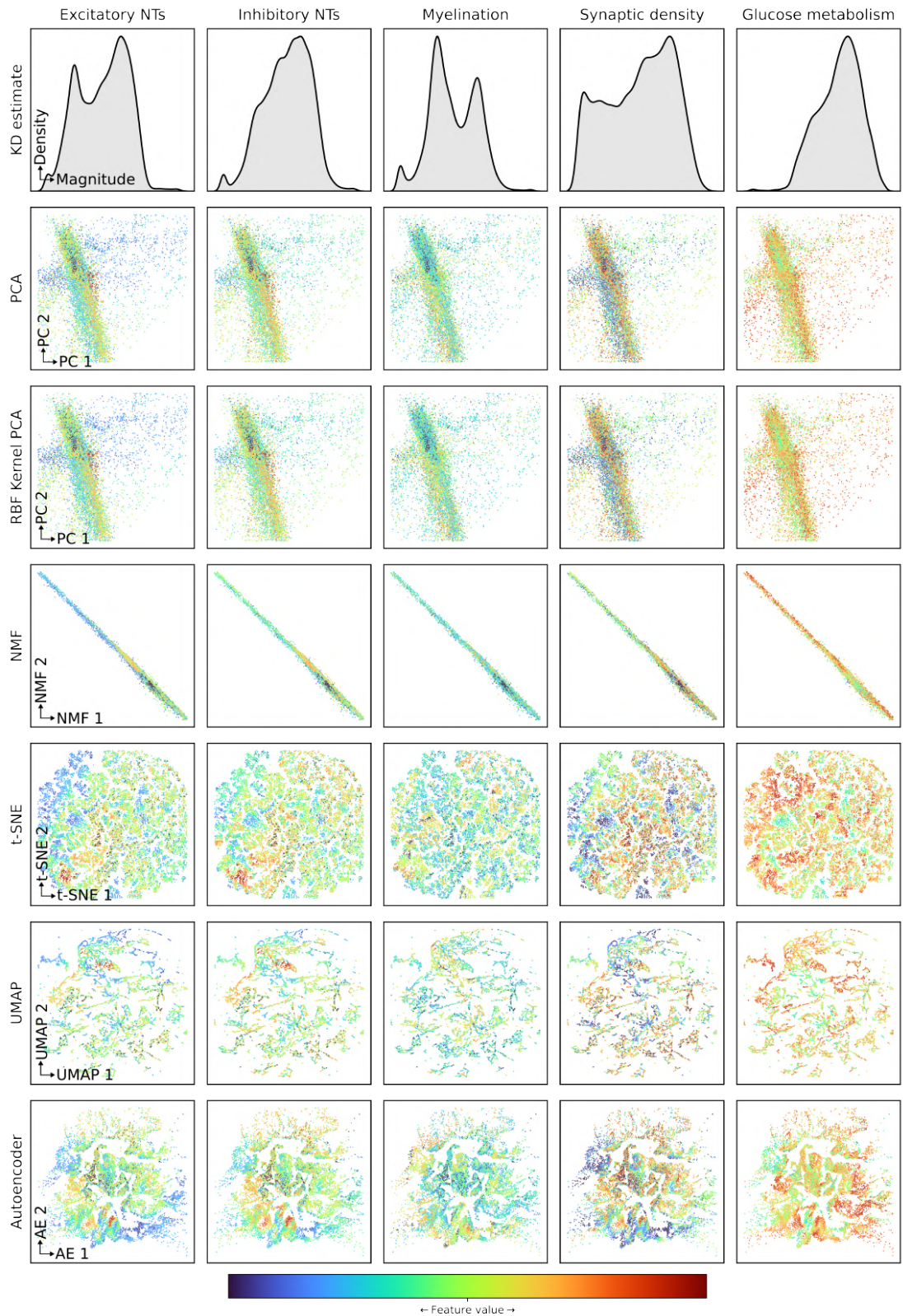

**Figure 4.** Feature annotation maps reveal the relationship between 2D transcriptional representations (rows) and brain signalling, microstructural, and metabolic characteristics (columns). Plots of each voxel from each representation are coloured by the feature value from the corresponding map, expressed in arbitrary units: A) mean value of the excitatory neurotransmitters (NTs), B) mean value of the inhibitory neurotransmitters, C) myelination, D) synaptic density, and E) glucose metabolism. The first row shows a kernel density (KD) estimate of each feature distribution.

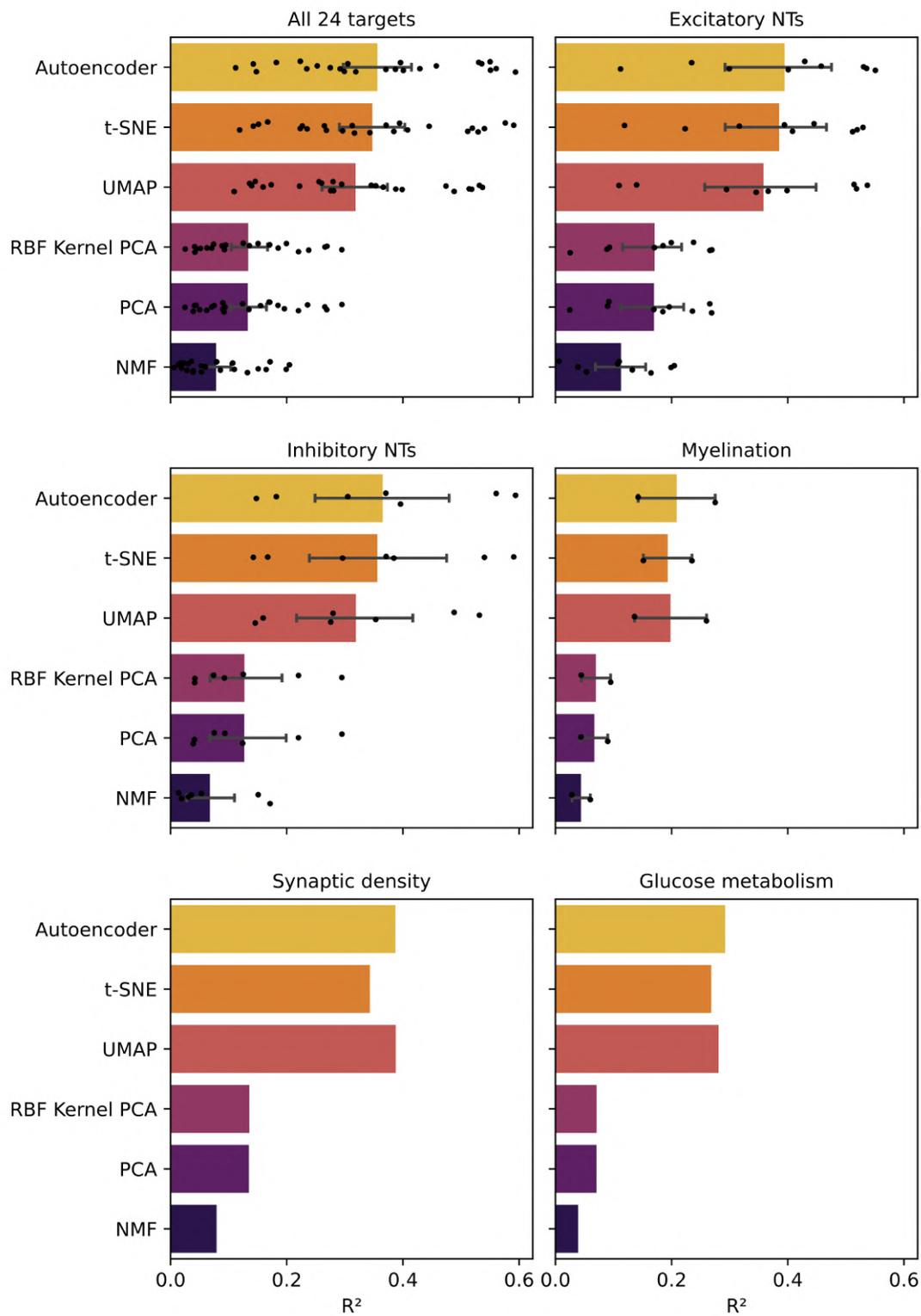

Figure 5. Prediction of brain signalling, microstructure, and metabolism from transcriptional representations. A) Prediction performance across all 24 modelling targets, B) all excitatory neurotransmitter (NT) targets (both individually and mean thereof), C) all inhibitory neurotransmitter targets (both individually and mean thereof), D) myelination targets, E) synaptic density, and F) glucose metabolism. The representation learning method is shown on y-axis, and test-set $R^2$ value along the x-axis. The bar colour is mapped to the mean test RMSE. Individual experiments are plotted as individual points, with error bars, where applicable, indicating ±1 standard deviation.

## Discussion

We have systematically evaluated the comparative characteristics of linear and non-linear compressed representations of high-resolution, whole-brain gene transcriptomic data, quantifying out-of-sample reconstruction fidelity and downstream predictive power across an array of signalling, microstructural, and metabolic targets. Our analysis seeks to provide definitive answers to two key questions concerning the anatomical modelling of brain transcription: first, does PCA—the dominant current approach—provide a plausibly adequate representation, and second, what form of representation achieves the optimum balance of fidelity and compression?

The answer to the first question is clear: the first two components of PCA capture comparatively little transcriptional variance, yield highly reductive spatial maps with minimal relation to underlying neuroanatomy, and are weakly predictive of underlying signalling, microstructural, and metabolic characteristics compared with the alternative methods at our disposal. If low-dimensional PCA representations omit most of the transcriptional signal, their capacity to contextualise other anatomically organised characteristics within mechanistic models of brain function and dysfunction must be correspondingly limited. In such circumstances, the form of organisation of any given latent component—for example, a smooth gradient—need not imply a corresponding form of native organisation. Smoothness, interpreted as a general organisational principle, may be merely an artefact of the representation. In general, the lower the fidelity of the representation, the higher the risk that its properties do not reflect the underlying data generative process. Consider, for example, how little of human craniofacial morphology is conveyed in the first two "eigenfaces"[59]. Our evaluation here is twofold. Firstly, the quality of a low-dimensional representation is commonly quantified by the amount of captured variance. Many users of PCA report the cumulative or individual variance of the reported components. A more general approach, applicable to non-linear methods, is to measure the error of the reconstructions from the latent representation. An ideal latent representation should yield a perfect reconstruction of the data; the larger the error, the worse the representation. Secondly, we can test the predictive power of the representation in downstream discriminative modelling, leaning on the reasonable expectation that whole brain transcription is related to neurotransmission, myelination, synaptic density, or metabolism. High predictive fidelity indicates a better (and more useful) representation. We show throughout these works the inferiority of PCA across all these tasks.

The answer to the second question is also clear: deep auto-encoders offer superior reconstruction fidelity and biological plausibility over other methods, while exhibiting predictive utility comparable with UMAP and t-SNE. This is especially prominent in the prediction of neurotransmitter receptor density – both excitatory and inhibitory – as well as synaptic density, with weaker but nonetheless present relation to myelination quantities and glucose metabolism. Note the real-world utility of t-SNE is limited by the absence of an explicit, reusable mapping function. Moreover, the architectural flexibility of auto-encoders enables customisation to diverse input data types, and through denoising[60], masked[61], hierarchical[62], and variational[63] variants, offers adaptability to noisy, missing, multi-scale, and heterogeneous data. Variational auto-encoders also facilitate the crafting of shaped representations that emphasise features of interest while minimising nuisance confounders[63-65]. Convolutional auto-encoders[66] enable graceful modelling of spatial dependencies within patch-wise (multi-voxel as opposed to single voxel) models, indeed provide a means of scaling to whole brain patterns when large-scale transcription data eventually become available. Latent representations are typically optimised for compactness while retaining individual variability and eliciting characteristic structure: disentanglement. However, it may also be desirable to emphasise differences of specific interest while minimising others, such as removing confounders in a representation, for example, irrelevant biological effects or non-randomness in treatment allocation. The current models use a single voxel, at a given location, as input, handling different voxels as replications. But an ideal model would take more than one voxel as the input, indeed ideally the entire brain. Replication is then not over voxels, but over brains, and necessarily requires thousands of brains to render statistically tractable.

This flexibility comes at a cost. The implementation and training of auto-encoders requires skill, and involves architectural and hyper-parameter choices[23], for which the available data may provide only limited guidance. The greater the model flexibility, the higher the risk of overfitting, though attention to training and evaluation can minimise it. Without large-scale replication across individuals, the analysis of transcription patterns unavoidably ignores variability across the population that may limit the generalisable detail of any single representation. But this is a problem only inter-individual replication could conceivably remedy, and the solution necessarily requires models of greater, not lesser, expressivity. In short, accessibility is the only defensible grounds for preferring PCA over deep auto-encoding in the derivation of compressed anatomically organised representations of whole brain transcription data.

## Conclusion

We systematically evaluate an array of dimensionality-reduction techniques applied to large-scale transcriptomic data across the whole human brain in 1) characterising a latent structure, 2) reconstructing original microarray transcriptomic data in held-out test sets, and 3) downstream prediction of a range of physiological data including local neurotransmitter expression, synaptic density, myelination, and metabolic activity. Across all three, we show that closer approximations of the source data and higher fidelity predictions are achievable using deep autoencoding over PCA, kPCA, UMAP, t-SNE, or NMF, findings supportive that the path to transcriptomic-neuroimaging innovation lie with the utilisation of sufficiently complex and nonlinearly capable high-dimensional representation methods. In making our model and code freely accessible, we hope to encourage researchers to use the representational method that currently best preserves the transcriptional signal across the human brain.

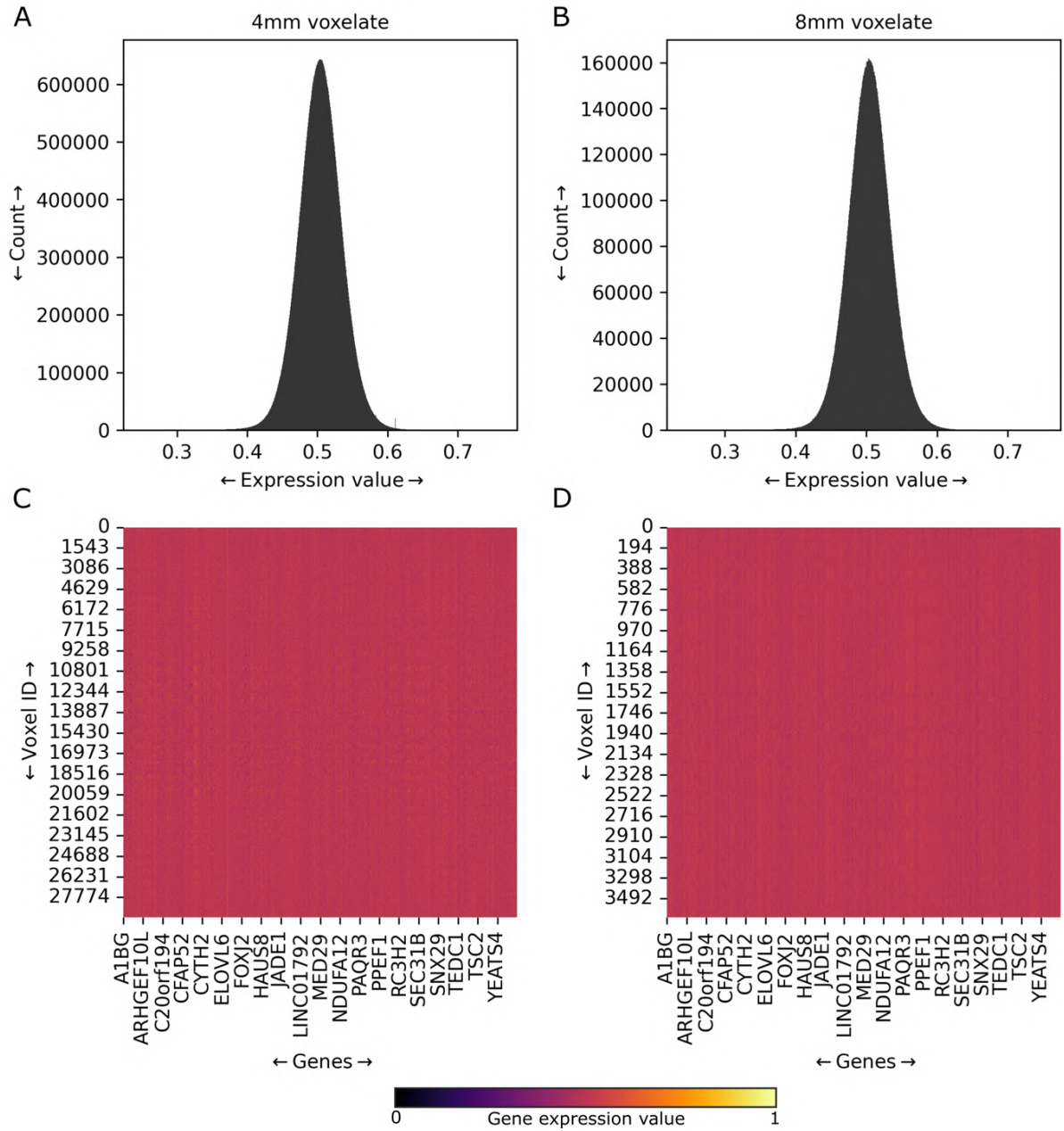

**Supplementary Figure 1.** Histograms of transcriptomic data at 4mm$^3$ (A) and 8mm$^3$ resolutions (B) shows the distribution to be Gaussian. Heatmaps in C) and D) show individual voxel x gene expression values in 4mm$^3$ and 8mm$^3$ resolutions, respectively.